\documentclass{article}

\usepackage{microtype}
\usepackage{graphicx}
\usepackage{subcaption}
\usepackage{booktabs} 

\usepackage{hyperref}

\usepackage[accepted]{icml2026}

\usepackage{amsmath}
\usepackage{amssymb}
\usepackage{mathtools}
\usepackage{amsthm}

\usepackage[capitalize,noabbrev]{cleveref}

\theoremstyle{plain}

\theoremstyle{definition}

\theoremstyle{remark}

\usepackage[textsize=tiny]{todonotes}

\icmltitlerunning{Steerable Cultural Preference Optimization of Reward Models}

\begin{document}

\twocolumn[
  \icmltitle{Steerable Cultural Preference Optimization of Reward Models}



  \icmlsetsymbol{equal}{*}

  \begin{icmlauthorlist}
    \icmlauthor{Minsik Oh}{sss}
    \icmlauthor{Advit Deepak}{sss}
    \icmlauthor{Sophie Wu}{sss}
    \icmlauthor{Douwe Kiela}{sss}
    \icmlauthor{Ekaterina Shutova}{aaa}
  \end{icmlauthorlist}

  \icmlaffiliation{sss}{Stanford University}
  \icmlaffiliation{aaa}{University of Amsterdam}

  \icmlcorrespondingauthor{Minsik Oh}{minsik@stanford.edu}

  \icmlkeywords{Machine Learning, ICML}

  \vskip 0.3in
]



\printAffiliationsAndNotice{}  

\begin{abstract}
It is essential for large language model (LLM) technology to serve many different cultural sub-communities in a manner that is acceptable to each community. However, research on LLM alignment has so far predominantly focused on predicting a unified response preference of annotators from certain regions. This paper aims to advance the development of alignment models with a more global outlook, that are able to accurately represent the preferences of subcommunities and do not exhibit excessive bias towards any of them. We focus on the development of reward models for this purpose and present a novel reward model training algorithm (SCPO) that can incorporate diverse cultural preferences in a balanced manner.
Our method results in performance increases of the minority reward model of up to 7 points over the baseline model across two datasets, PRISM and GlobalOpinionQA, and across 7 countries. SCPO is up to 280\% more training data-efficient than full-data finetuning of reward models.
In addition, we perform analysis of bias by separately evaluating on the preference of subcommunities and show that excessive bias is mitigated via our weighting method. Our code is available at \url{https://github.com/minsik-ai/Steerable-Cultural-Preference}.
\end{abstract}


\section{Introduction}

\begin{figure*}[t]
\includegraphics[width=1.4\columnwidth]{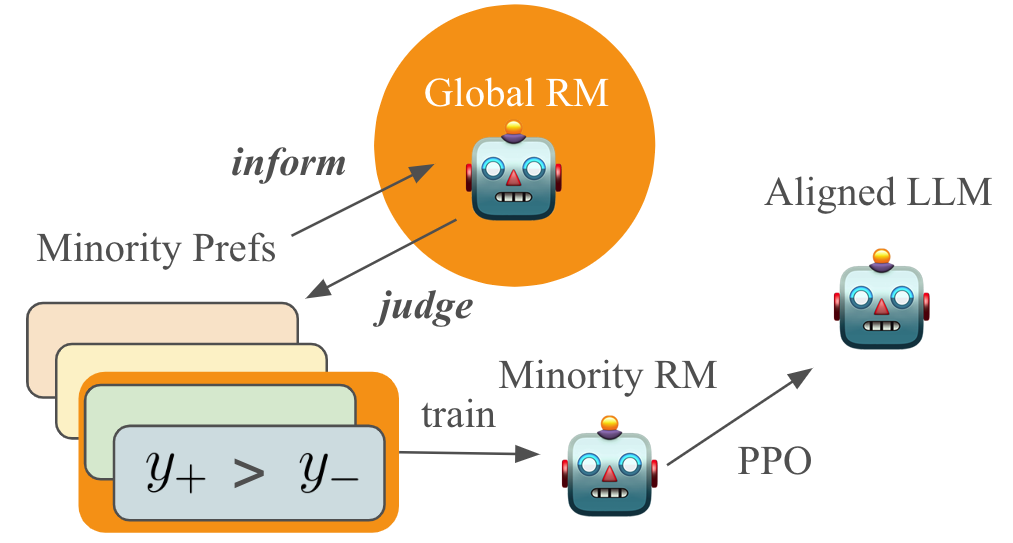}
\centering
\caption{Overview of Steerable Cultural Preference Optimization (SCPO). The pipeline consists of: (1) a global reward model that scores minority preference pairs; (2) \textbf{Filtering} (Section~\ref{filtering}): removing pairs where minority and global preferences agree; and (3) \textbf{Weighting} (Section~\ref{weighted}): assigning lower training weights to divergent preferences. Orange highlighting indicates our novel contributions. The right side shows downstream use in PPO-based RLHF. Note that the global reward model may be same as starting checkpoint of RM for minority training.}
\label{fig_algo}
\end{figure*}


Aligning large language models (LLMs) to individual group (minority) preferences is an important open problem~\citep{gpo} that has seen measured progress on demographic and country-specific evaluations~\citep{opqa, durmus2024measuringrepresentationsubjectiveglobal}. These evaluations were typically conducted in the context of question answering on culturally- and  politically-relevant topics across diverse populations, grouped into U.S. states and other demographic factors~\citep{opqa} and distinct countries~\citep{durmus2024measuringrepresentationsubjectiveglobal}. LLMs are known to reflect opinions from either privileged populations~\citep{opqa} or over-representing opinions from Western, developed countries~\citep{durmus2024measuringrepresentationsubjectiveglobal}, making minority-aligned language modelling an urgent problem.

Minority alignment is a problem defined under the umbrella of pluralistic alignment~\citep{pluralistic}. Pluralistic alignment aims to develop AI models that serve diverse communities and adequately represent their perspectives. \citet{pluralistic} proposed three types of pluralistic alignment: overton, where the model outputs diverse perspectives; steerable, where the model can be steered to output a particular perspective; and distributional, where a distribution of perspectives is modelled explicitly. Our approach to minority alignment aims to build steerable reward models that are specific to a country's point of view.

Several recent alignment frameworks aim to model group preferences. These include methods such as Group Preference Optimization (GPO)~\citep{gpo} and Group Robust Preference Optimization (GRPO)~\citep{grpo} can train a group preference model. GPO utilizes a separate fine-tuned transformer module on top of LLM to predict a group's preferences. This makes it not straightforward to integrate into general-purpose LLM alignment frameworks, such as reinforcement learning with human feedback (RLHF)~\citep{ppo} or direct preference optimization (DPO)~\citep{dpo}, as it has not been developed with this in mind. GRPO, on the other hand, works with a specific definition of ``robustness" and minimizes the worst-case group loss. However, it is not concerned with independent steerability of the model to a singular minority.

Central to our approach is the use of a 'global' reward model: an RM trained on broad preference data (e.g., OpenAssistant or Tülu 3). We leverage this global RM for two purposes:

\begin{enumerate}
    \item \textbf{Identifying Cultural Distinctiveness}: Preference pairs where minority 
annotations disagree with global RM predictions represent genuinely distinctive 
cultural preferences. By filtering to retain only these disagreeing pairs, we 
focus training on what makes each culture unique rather than on universal 
preferences already captured by the global model.
    \item \textbf{Measuring Divergence}: Large disagreements (where the global RM assigns 
very different scores than minority annotations) may indicate divergent preferences 
that risk making the minority model overly biased. Our weighting scheme 
down-weights such divergent cases.
\end{enumerate}

The global RM thus serves as a reference point representing 'mainstream' or 'consensus' preferences, enabling us to systematically identify and calibrate cultural deviations. Importantly, we do not assume the global RM is 'correct'. Rather, we use it as a tool to differentiate minority preferences from majority ones (examples in Fig.~\ref{fig_example}).

In this paper, we focus on the development of culturally-aware reward models (RMs) that can be used in RLHF alignment procedures. Specifically, we propose a novel method that utilizes a global reward model to identify culture-specific preference samples and present a weighted reward model training loss to conduct a multi-faceted balanced training of RMs.  Our research questions are as follows:

\begin{enumerate}
    \item \textbf{How do we ensure that minority reward models have balanced opinions?} While we want to reflect minorities' opinions on LLM outputs, we want to simultaneously de-emphasize undesired responses within a minority preference dataset. We design a two-tiered multi-faceted evaluation approach that utilizes distinct test sets to ensure we create a reward model with balanced opinions.

    \item \textbf{Can we utilize global reward model preference scores for minority reward model training?} We devise a novel alignment method that utilizes open-source reward models that are not minority aligned. We utilize the scores given by these global reward models for both training and evaluation of the minority reward models.

    \item \textbf{Which subsection of preference data is important for effective minority reward model training?} Some training preference pairs in minority preference data will agree with the global model, while other pairs will be different. We utilize the scores of the global reward models on certain preference pairs to either truncate or emphasize sections of pairwise preference data, and report observed performance tradeoffs.
\end{enumerate}

We fine-tune two reward models (OpenAssistant and Tulu) on country-specific data from the PRISM dataset using our method and find filtering and weighting of the data, utilizing global reward model scores, is beneficial to the performance of our models on overall test set, while avoiding aligning to skewed preference.

\section{Related Works}

Prior cultural alignment work has explored prompting~\citep{li2024culturegen, investigating, CultureLLM}, but these rely on crafted prompts without real preference data. Approaches such as reward ranked fine-tuning (RAFT) ~\citep{dong2023raft} and Supervised Iterative Learning from Human Feedback (SuperHF) ~\citep{mukobi2023superhf}, demonstrate the potential of using only the most valuable training examples to improve model performance. RAFT utilizes reward-based reranking by iteratively scoring samples via a reward function, filtering for high-reward examples, and fine-tuning the model using this subset. Similarly, SuperHF filters model-generated training data with a reward model and only uses high-reward synthetic data for fine-tuning. Both approaches demonstrate significant improvements by using a reward model to identify high-quality data. However, neither method targets minority alignment, accounts for preference pairs, or goes beyond basic reward thresholds for filtering.

Methods in weighting-based alignment, such as Online Preference Tuning (OPTune) ~\citep{chen2024doptune} and Mallows-DPO ~\citep{chen2024bmallowsdpo}, highlight the benefits of using reward models to prioritize certain samples. OPTune improves alignment by introducing a weighted DPO objective that emphasizes pairs with larger reward gaps, ensuring the model learns more from high-priority examples. Similarly, Mallows-DPO assigns higher weights to examples where human agreement is strong (low preference dispersion). Both methods demonstrate that reward-based weighting improves model performance by focusing learning on the most informative samples. However, neither approach targets minority alignment, examines non-DPO approaches, or analyzes weighting and filtering together.

A parallel body of work on personalized reward modeling exists. PAL~\citep{pal} represents each user as a weighted combination of learnable preference prototypes, enabling few-shot adaptation to unseen users; VPL~\citep{vpl} encodes user-specific preferences into a latent variable via a variational autoencoder. The Multilingual Alignment Prism~\cite{multiharm} balances optimizing for multilinguality against minimizing global and local harms. SCPO differs in two ways. First, rather than learning a user-conditioned model, it operates on population-group identity through a global-versus-minority contrast, using an off-the-shelf global RM as a fixed reference. This enables a drop-in integration with standard RLHF without an auxiliary preference module. Second, these methods are largely evaluated on synthetic or image-preference benchmarks (e.g., PAL on Pick-a-Pic, VPL on Pets) and report aggregate performance over seen/unseen users rather than per-population statistics, whereas our analysis is grounded in human per-country data and reports the minority-versus-overall trade-off per population. Since per-user and per-population alignment are distinct objectives, user-level alignment does not establish balanced group-level alignment, motivating the population-stratified approach we pursue.

\section{Datasets}
\label{data}

\paragraph{PRISM} We primarily utilize PRISM~\citep{prism}, a human feedback dataset for preference and value alignment of LLMs. PRISM is an LLM preference dataset comprising of controversial conversations between LLM and user across different countries. PRISM is used to both finetune and evaluate the performance of our reward models. We randomly split PRISM users into train and test sets using 8.5:1.5 user ratio (stratified per country), to ensure multi-turn data from conversations are not divided across the data splits. Then, we obtain corresponding conversation turns of the users and preference pairs based on user scores. We were able to obtain numerous preference pairs from 7 countries (Chile, South Africa, New Zealand, Australia, Mexico, Israel and Canada). 

To fit our use cases, we re-structure both the \texttt{survey} data (which contains demographic information of the participants, as seen in Appendix~\ref{appendix:data} Table \ref{table:prism_survey}) and the \texttt{utterance} data (the content of the actual conversations between participants and LLMs and participant ratings, as seen in Appendix~\ref{appendix:data} Table \ref{table:prism_utters}) from PRISM. 

\paragraph{GlobalOpinionQA} Additionally, we use Anthropic's GlobalOpinionQA ~\citep{durmus2024measuringrepresentationsubjectiveglobal} dataset to evaluate our country-specific reward models. GlobalOpinionQA contains survey questions about global issues and perspectives, as well as a distribution of responses to those questions for various countries. By providing the question as the prompt and each of the answer options as responses to the country-specific reward model, we can see if the relative ranking of the rewards given to each answer corresponds with the probability distribution of answers chosen by that country in GlobalOpinionQA.



\section{Methodology}
\label{our_methods}

\begin{figure*}[t]
\includegraphics[width=2\columnwidth]{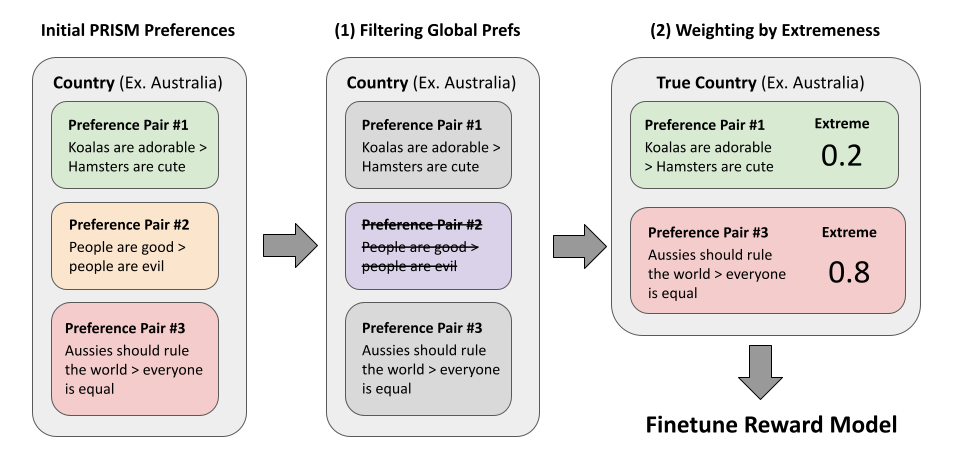}
\centering
\caption{Detailed diagram of our filtering and weighting method. The first step is retrieving all country-specific PRISM preferences. Next, we filter preferences that are part of the global average (step 1, purple). Then, we identify the divergence of each of the true country preferences (step 2, green is benign, red is divergent). Finally, we use this weighted subset to finetune the reward model. See Section~\ref{our_methods} for details. Samples in Appendix~\ref{app:samples} Table~\ref{table:extremeness_good},~\ref{table:extremeness_bad}.}
\label{fig_example}
\end{figure*}

We develop a novel method (Fig.~\ref{fig_algo}) of working with minority preferences in conjunction with global preferences, which consist of filtering and weighting stages.  Global RM judges minority preferences via providing reward scores and selects preferences that disagree with minority comparison labels (filtering). Using the Global RM reward scores, each preference is weighted differently in weighted training loss, to ensure subtle differences are emphasized (weighting). Global RM can be reused from starting Tülu 3 and OpenAssistant models, while minority RM is a result of training the said models to given minority's preferences. See Fig.~\ref{fig_example} for an example.

\subsection{Filtering}
\label{filtering}

We remove minority pairwise preferences from the training set if they agree with the global model preferences. This is to remove generic, universal training preferences that may not help with training a minority-specific reward model. Conversely, we retain preference pairs that disagree with the global model. By keeping only the minority pairwise preferences that disagree with the global model preferences, we aim to streamline the training of minority reward models by utilizing only necessary data to achieve greater data efficiency. In practice, about one half or one third of training data is left after filtering, achieving 170\% - 280\% data efficiency. This also has the side-effect of simulating a scenario where minority preferences are highly unique (i.e. 90\% of the preferences disagree with the global consensus).

We utilize the Bradley-Terry model~\citep{btmodel} for our filtering algorithm. We retain preference pairs where the Global RM disagrees with the minority label (indicating a unique cultural preference) and discard pairs where the Global RM already aligns with the minority label (indicating a generic preference). Our filtering algorithm is as follows:
\begin{equation}
p_{\text{glo}}(y_+ \succ y_- | x) = \frac{e^{r_{\text{glo}}(x, y_+)}}{e^{r_{\text{glo}}(x, y_+)} + e^{r_{\text{glo}}(x, y_-)}} < \tau
\label{eq_filter}
\end{equation}
Per minority preference annotations, $y_+$ is the preferred response in the pair and $y_-$ is the dispreferred response in the pair. $p_{\text{glo}}(y_+ \succ y_- | x)$ is the probability that corresponds to the global model preferring $y_+$ data instead of $y_-$. $r_{\text{glo}}$ is global reward model that produces a score. $\tau$ is a $0 \leq \tau \leq 1$ threshold for subset selection of preference data, and lower value indicates more aggressive filtering.

\subsection{Weighted RM training loss}
\label{weighted}
We define the 'divergence' of a preference pair $(y_{+}, y_{-})$ as the degree of disagreement between the minority annotation and the global RM's preference. Formally, a preference is divergent when $p_{\text{glo}}(y_+ \succ y_- | x)$ is high, meaning the global model strongly prefers the response that minority annotators rejected. We distinguish divergence from related concepts:

\begin{itemize}
\item \textbf{Harmful content} - Responses containing toxic language, hate speech, or unsafe recommendations. While divergent preferences may correlate with harmful content (see Table 2), divergence is defined purely by disagreement magnitude, not content analysis. 

\item \textbf{Cultural distinctiveness} - Preferences can be culturally distinctive without being divergent—e.g., preferring formal vs. informal language styles.

\item \textbf{Annotator error} - Some divergent preferences may reflect inconsistent annotations rather than genuine cultural differences.
\end{itemize}

Our weighting scheme does not classify content as 'good' or 'bad' but rather modulates training influence based on disagreement magnitude, allowing the model to learn from distinctive preferences while reducing influence of outliers.

We develop a novel training loss that inversely assigns weights to the preferences pairs according to their divergence. Thus, less divergent preference data is weighted more highly than more divergent preference data. (See Table~\ref{table:extremeness_good} and Table~\ref{table:extremeness_bad} for examples of responses and their associated divergence.) With this approach, we aim to ensure that more divergent characteristics of minority preferences are dulled in favor of subtle, important cultural differences that make the minorities unique. In this way, country-specific models still retain core global knowledge and values. Our new training loss (Eq.~\ref{weigh_loss}) utilizes the global reward model reward scores to determine the weights per preference pairs.

Our weighting scheme builds on the Bradley-Terry model~\citep{btmodel}. A key insight from this framework is that preference annotations are inherently probabilistic. When two responses have similar quality, the preference probability is close to 0.5 and observed annotations are noisy. When responses differ substantially, preferences become more deterministic.

We extend this view to the relationship between global and minority preferences. Consider a preference pair where the minority annotation indicates $y^+ \succ y^-$ but the global RM assigns $p_{\text{glo}}(y^+ \succ y^- | x) < 0.5$. The disagreement magnitude admits two interpretations:

\begin{enumerate}
    \item \textbf{Cultural signal strength} - Small disagreements (global probability close to 0.5) indicate preferences where global and minority populations are nearly indifferent: the minority preference represents a subtle but genuine cultural distinction. Large disagreements (global probability close to 0) suggest the minority annotation contradicts strong global consensus.
    
    \item \textbf{Annotation reliability} - Under a mixture model where minority annotations arise from either (a) genuine cultural preferences or (b) noise/error, the posterior probability of genuine cultural signal decreases as disagreement magnitude increases.
\end{enumerate}

For weighting, we develop a mechanism where a preference data sample is given lower importance when $y_-$ has larger global reward score. Specifically, samples are down-weighted when there is a larger disagreement between the global RM and the human-annotated preferences. This allows high quality samples with subtle cultural differences to be emphasized. We define weight $W(y_+, y_-)$ as follows:
\begin{equation}
\begin{split}
    W(y_+, y_-) & = \min\left(\frac{1}{p_{\text{glo}}(y_- \succ y_+ | x)} - 1, 1\right)^{1/\beta} \\ & =
    \min\left(\frac{e^{r_{\text{glo}}(x, y_+)}}{e^{r_{\text{glo}}(x, y_-)}}, 1\right)^{1/\beta} \\ & =
    \min\left(e^{(r_{\text{glo}}(x, y_+) - r_{\text{glo}}(x, y_-))/\beta}, 1\right)
\end{split}
    \label{eq_weigh}
\end{equation}
$y_+$, $y_-$ and $r_{\text{glo}}$ are defined as in Eq.~\ref{eq_filter}. $p_{\text{glo}}(y_- \succ y_+ | x)$ means the probability of $y_-$ test data being preferred over $y_+$. The temperature hyperparameter $\beta > 0$ controls the sharpness of the weighting: smaller values of $\beta$ amplify the distinction between high and low-confidence pairs, while larger values yield more uniform weights. Note that for weight $<1$, $r_{\text{glo}}(x, y_-) > r_{\text{glo}}(x, y_+)$ which means the disagreement exists between the global model and the human-annotated preferences for this preference data.

We utilize the binary ranking loss to train our reward models, defined as follows:
\begin{equation}
    L = - \mathbb{E}_{(x, y^+, y^-)\sim D} [\log \sigma(r(x, y_+) - r(x, y_-))]
\label{base_loss}
\end{equation}
With preference data $(x, y^+, y^-) \in D$ where $y^+$ is preferred over $y^-$ for prompt $x$. $r$ is the reward function of an RM. 

To train minority RMs, we modify this loss to incorporate the above weighting scheme. Eq.~\ref{base_loss} becomes (with slight simplification of notation):
\begin{equation}
    L = - \mathbb{E}_{D} [W(y_+, y_-) \log \sigma(r(x, y_+) - r(x, y_-))]
\label{weigh_loss}
\end{equation}
Note that $r$ is the reward model to be trained, and differs from $r_{\text{glo}}$.

\section{Experimental Setup}

\subsection{Models and Training}

\begin{table*}[t]
\centering
\begin{tabular}{lrrrrrrrr}
\toprule
    & Chile & S. A. & N. Z. & Aus. & Mex. & Israel & Can. & Avg. \\ \midrule
Global RM & 54.54  & 64.06  & 56.55 & 59.61 & 51.64 & 63.03 & 60.40  & 58.55 \\ 
Baseline & 60.03 & 61.80  & \textbf{62.58} & 59.93 & 60.93 & 65.96 & 63.58 & 62.12 \\
Filtered Only & 51.59  & 39.62  & 52.83 & 41.77 & 52.88 & 39.67 & 49.71 & 46.87 \\
Inverse Weighted Only & 60.03 & 50.77 & 60.72 & 47.53 & 60.35 & 55.99 & 60.98 & 56.62 \\
SCPO (W) & 58.94 & 64.77 & 58.96 & 60.18 & 56.80 & 65.61 & 62.62 & 61.13 \\
SCPO (F + W) & \textbf{61.11} & 60.38 & 61.93  & 59.20 & \textbf{67.65} & 64.32 & 64.45 & 62.72 \\
SCPO (F + W)$_{\text{tuned}}$ & 59.89 & \textbf{64.17} & 62.30 & \textbf{60.26} & 63.39 & \textbf{67.84} & \textbf{66.09} & \textbf{63.42} \\
\bottomrule
\end{tabular}
\caption{Evaluations of methods using OpenAssistant RM, evaluating on all country-specific PRISM preferences. Bold indicates the best-performing method per column. See Section~\ref{exp_overall} for analysis. See Table~\ref{table:prism_all_w_errors},~\ref{table:prism_all_w_errors_2} for detailed results and error bars.}
\label{table:prism_all}
\end{table*}
\begin{table*}[h]
\centering
\begin{tabular}{lrrrrrrrr}
\toprule
      & Chile & S. A. & N. Z. & Aus. & Mex. & Israel & Can. & Avg. \\ \midrule
Baseline & 25.55 & 25.74 & 30.32 & 24.50 & 34.46 & 22.54 & 28.47 & 27.37 \\
Filtered Only & 57.94 & 61.72 & 58.71 & 59.24 & 70.62 & 73.65 & 59.18 & 63.01 \\
Inverse Weighted Only & 43.01 & 49.50 & 44.52 & 47.99 & 56.50 & 40.00 & 45.35 & 46.70 \\
SCPO (W) & 16.83 & 17.82 & 16.56 & 14.66 & 20.34 & 19.05 & 22.21 & 18.21 \\
SCPO (F + W) & 36.98 & 37.29 & 44.95 & 33.33 & 54.80 & 38.10 & 38.57 & 40.57 \\
SCPO (F + W)$_{\text{tuned}}$ & 28.10 & 27.39 & 28.82 & 22.69 & 40.68 & 25.40 & 27.74 & 28.69 \\
\bottomrule
\end{tabular}
\caption{Evaluation of methods using OpenAssistant RM, evaluating on true country-specific PRISM preferences. Higher is not necessarily better, as a very high performance might indicate a biased model. See Section~\ref{exp_true_openassistant}, Figure~\ref{tulu_tradeoff} for analysis. See Table~\ref{table:prism_true_w_errors},~\ref{table:prism_true_w_errors_2} for detailed results and error bars.}
\label{table:prism_true}
\end{table*}

\begin{table*}[t]
\centering
\setlength{\tabcolsep}{4pt} 
\begin{tabular}{lrrrrrrrr}
\toprule
 & Chile & S. A. & N. Z. & Aus. & Mex. & Israel & Can. & Avg. \\
\midrule
Global RM & 63.64  & 63.35  & 61.56 & 66.18 & 51.64 & 62.68 & 68.79 & 62.55\\
Baseline & 63.64  & 61.45  & \textbf{65.65} & 65.04 & 52.19 & 62.54 & \textbf{69.55} & 62.86\\
Filtered Only & 36.65 & 35.83 & 43.55 & 35.85 & 51.91 & 33.57 & 36.42 & 39.11 \\
Inverse Weighted Only & 63.85 & 61.21 & 62.49 & 63.91 & 53.28 & 61.85 & 65.89 & 61.78\\
SCPO (W) & 63.64  & \textbf{63.70} & 61.84 & \textbf{66.58} & 52.73 & \textbf{65.49} & 67.05 & \textbf{63.01}\\
SCPO (F + W) & \textbf{64.07} & 62.51 & 61.09 & 64.80 & 53.01 & 61.74 & 62.33 & 61.36\\
SCPO (F + W)$_{\text{tuned}}$ & 63.28 & 63.35 & 61.00 & 65.61 & \textbf{53.55} & 62.44 & 63.30 & 61.79 \\
\bottomrule
\end{tabular}
\caption{Evaluation of methods using Tülu3 RM, evaluating on all country-specific PRISM preferences. Bold is best method per country. See Section~\ref{exp_tulu3} for analysis. See Table~\ref{table:prism_tulu_w_errors},~\ref{table:prism_tulu_w_errors_2} for detailed results and error bars.}
\label{table:tulu}
\end{table*}





\begin{table*}[t]
\centering
\setlength{\tabcolsep}{4pt} 
\begin{tabular}{lccccc}
\toprule
 & Chile & Australia & Mexico &  Canada & Avg. \\
\midrule
Global RM & 83.04 & 82.10 & 83.97 & 82.85 & 82.99 \\
GPO & 83.16 & \textbf{82.78} & 83.42 & 83.73 & 83.27 \\
SCPO & \textbf{92.57}  & 81.76 & \textbf{92.53} & \textbf{91.87} & \textbf{89.68} \\
\bottomrule
\end{tabular}
\caption{Evaluation of best-performing OpenAssistant SCPO and GPO methods from GlobalOpinionQA. Bold indicates highest value. Only countries where we have best results from SCPO in Table~\ref{table:prism_all} are shown. See Section~\ref{gqa_tulu} for analysis and country selection process.}
\label{table:results_gqa}
\end{table*}

\paragraph{Training data} We train our RMs using our split of PRISM training set from 7 countries (Chile, South Africa, New Zealand, Australia, Mexico, Israel and Canada). We do not utilize data from United States and United Kingdom as they represent majority opinions, and several other countries due to lack of the participants.

\paragraph{Reward models} We utilize Tülu-3-8B\footnote{allenai/Llama-3.1-Tulu-3-8B-RM in HuggingFace} RM~\citep{tulu3} and OpenAssistant DeBERTa-V3-base\footnote{OpenAssistant/reward-model-deberta-v3-base in HuggingFace} RM~\citep{deberta1, deberta2}. 
See Appendix~\ref{app:hyper} for hyperparameters. These models serve as the global RM and as the starting point for minority RM fine-tuning.

\paragraph{Baselines} For each country $X$, we evaluate the following methods:
\begin{itemize}
    \item \textbf{Global RM} - Directly use the global RM, can be same as starting RM.

    \item \textbf{Baseline} - Fine-tune the global RM using all country $X$ PRISM preferences.

    \item \textbf{Filtered only} - Select country $X$ preferences from PRISM using the global RM and our filtering equation; Fine-tune the global RM using this subset of country $X$ preferences (i.e. country $X$-specific preferences)

    \item \textbf{Inverse weighting} We also experiment with inverse weighting method as another baseline (Appendix~\ref{app:inverse} Eq.~\ref{eq_inv_weigh}) that emphasizes disagreement instead of dampening it.
\end{itemize}



\subsection{Evaluation}
For our overall evaluations, we utilize the full PRISM test set for each country. In addition, we create a new minority-centric subset of the test set to ensure that we obtain a holistic overview of the minority RM's performance in regards to the divergence of minority opinions. It may be possible that a minority RM would align disproportionately to the more divergent preferences that are available in the minority dataset, losing alignment performance on global preferences. To measure this side-effect, we only collect minority preference pairs that are not consistent with the global model judgments and test whether the performance on this additionally selected test set is substantially higher (Fig.~\ref{tulu_tradeoff}).\footnote{For example, let sentences A \& B be 2 sentences in a preference pair. The condition for membership into the true country-specific subset is if country preference label says A \textgreater  B but the global model rewards says A \textless  B.}

We refer to these pairs as ``true country-specific subsets" of minority preferences and evaluate on them to identify reasons for overall performance changes. We do not claim these represent essential cultural preferences - rather, they operationally define the subset of preferences that differ from global consensus. This operationalization allows quantitative analysis but should not be interpreted as capturing authentic cultural values, which would require ethnographic validation beyond the scope of this work.

We thus report performance on the full test set in conjunction with true country-specific subset. We compute with 2 accuracy scores per country, one for full test set and another for true country-specific subset (Tables~\ref{table:prism_all} and~\ref{table:prism_true}, respectively). Higher performance on full test set means better performance of the RM, while performance on true country-specific subset should be analyzed in a nuanced manner, since having a high performance on this subset and low performance on full test set might indicate that the model is inappropriately skewed towards divergent and biased opinions.

\section{Experiments and Results}

\subsection{PRISM Experiments} \label{exp_openassistant}

We use the OpenAssistant RM to benchmark all methods on both PRISM and True-Country PRISM evaluations across seven countries (Tables~\ref{table:prism_all} and ~\ref{table:prism_true}). We omit U.S. and U.K. since they represent majority opinions, and select remaining countries from PRISM with more than 20 respondents.

\subsubsection{Overall Country Evaluation}
\label{exp_overall}

Starting with Table~\ref{table:prism_all}, we observe that the baseline outperforms the global RM, which can be expected as the baseline is the global RM fine-tuned on the country-specific preferences. Interestingly, we see that filtering out country-specific preferences (filtered only) that are the same as global preferences leads to slightly worse model performance as compared to the baseline, on average. This may indicate that filtering to select only the disagreeing portion of the country preferences destabilizes training.

We see that SCPO (either its weighted (W) or filtered \& weighted (F + W) variant) outperforms fine-tuning with all country-specific data, for 6 out of 7 countries. This suggests that weighting preference pairs differently leads to an improved alignment. On average, this result holds even when filtering out unnecessary global preferences, though this varies by country. Filtering is important in that it can increase the sample efficiency of training data. One caveat is that filtering only might have a negative effect of aligning the model too closely to true country specific preferences (as seen in Table~\ref{table:prism_true}), which may lead to poorer generalization to overall preferences expressed in the training data. Thus the combination of filtering with weighting (Section~\ref{weighted}) is critical, in that it helps the model to pay attention to subtle differences during training.


\subsubsection{True Country-Specific Evaluation}
\label{exp_true_openassistant}

Examining Table~\ref{table:prism_true}, we can see the results of our method on only the subset of true country-specific preferences. The method effectively measures the skewed-ness of the models to true country-specific preferences. Intuitively, SCPO (W) and SCPO (F + W) should have a lower score than using Filtering only model since we weight the importance of the samples such that skewed samples have less weight. We convincingly see this trend across all countries (on average, $-22.44$). This indicates that the weighting step is critical to balanced minority alignment, retaining the global preference signal (core values) while adopting non-divergent minority preferences.


\subsubsection{Tülu3 Experiments} \label{exp_tulu3}

Next, we apply our methods to a recent reward model, Tülu3-8B~\citep{tulu3}. We benchmark our method against the baselines as shown in Table~\ref{table:tulu}. We observe that the weighted loss we proposed in SCPO yields the best quality of alignment for most countries, as well as on average. Whilst in general the trends we observed are similar to those in case of the OpenAssistant model (Table~\ref{table:prism_all}), the filtering component of SCPO appears less useful for Tülu3 when hyperparameters are not tuned. This may be because the larger size of Tülu 3 models may lead to overfitting when trained on fewer, filtered preferences, requiring more thorough exploration. 


\subsubsection{Performance Tradeoff} \label{perf_tradeoff}

\begin{table*}[t]
\centering
\begin{tabular}{lcrrrrrrrr}
\toprule
    & Retained & Chile & S. A. & N. Z. & Aus. & Mex. & Israel & Can. & Avg. \\ \midrule
Global RM & 100\% & 54.54  & 64.06  & 56.55 & 59.61 & 51.64 & 63.03 & 60.40  & 58.55 \\
Weighted Only & 100\% & 58.94 & 64.77 & 58.96 & 60.18 & 56.80 & 65.61 & 62.62 & 61.13 \\
Random Filtering & 58.45\% & 54.55 & \textbf{65.84} & 60.17 & 58.64 & 56.56 & 63.38 & 62.72 & 60.26 \\
SCPO Filtering & 58.45\% & \textbf{60.61} & 64.41 & \textbf{61.56} & \textbf{60.58} & \textbf{64.75} & \textbf{68.31} & \textbf{65.90} & \textbf{63.73} \\
\bottomrule
\end{tabular}
\caption{Comparison of GlobalRM-informed filtering (SCPO) vs.\ random size-matched filtering using OpenAssistant RM, evaluating on all-country specific PRISM preferences. Bold is best method. Filtering methods are weighted per same configuration as weighted only. See Section~\ref{random_filtering} for analysis.}
\label{table:random_filtering}
\end{table*}

\begin{figure}[t]
\includegraphics[width=1.0\columnwidth]{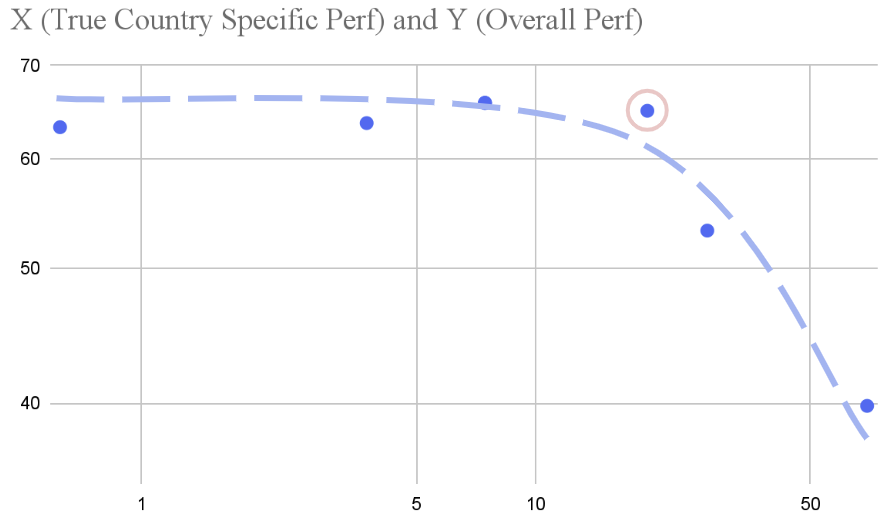}
\centering
\caption{Log-log graph trade-off of true-country (x-axis) vs. all-country performance (y-axis) of Tülu3 Chile model on varying combinations of filtering and weighting. Circled red is the optimal model. See Section~\ref{perf_tradeoff} for analysis.}
\label{tulu_tradeoff}
\end{figure}

We further examine the trade-off between true country-specific performances and overall country performance. We take the Tülu3 RM and vary SCPO's combination of filtering and weighting methods and their hyperparameters (learning rate) to produce six different finetuned Tülu3 RMs. We benchmark these RMs on PRISM's all-Chile preferences and true-Chile preferences to analye the trade-off (Figure~\ref{tulu_tradeoff}).

We can see a trade-off where filtering yields low performance for the overall country but high true-country performance, which matches our intuition that skewed samples from filtering may cause overfitting, from Table~\ref{table:prism_true}. Thus, our goal should be to align the models to have both high overall performance and true country-specific performance.

\subsection{GlobalOpinionQA Evaluation}
\label{gqa_tulu}

We further evaluate our RMs on GlobalOpinionQA (GQA) for the countries of Chile, Australia, Mexico and Canada. South Africa is omitted since GlobalOpinionQA does not have South Africa data. New Zealand and Israel are omitted since baseline models outperform SCPO models (Table~\ref{table:prism_all}). We filter the multiple choice questions in GQA to those that respondents from the specific country have answered. For each question, we pass each (question, option) pair through the baseline and SCPO RMs to get a score. We then compare these reward scores per given option and the ground truth percentages of respondents from the specific country who selected a given option (Table~\ref{table:results_gqa}). Specifically, we compute the Jensen-Shannon Distance (JSD) between these two distributions~\citep{durmus2024measuringrepresentationsubjectiveglobal} and use $1-JSD$ as our metric, indicating similarity of the RM scores with human responses. We also compare our method to the group preference optimisation (GPO) approach of \citet{gpo}. This results demonstrated that our SCPO method leads to a better cultural alignment than both the baseline and GPO.

\subsection{Ablation Analysis}
\label{ablations}

To validate the design choices of SCPO, we conduct ablation studies addressing three key questions: (1) whether improvements stem from GlobalRM-informed selection or simply data reduction (Section~\ref{random_filtering}), (2) sensitivity to the filtering threshold $\tau$ (Appendix~\ref{sec:ablation-tau}) and (3) sensitivity to the weighting threshold $\beta$ (Appendix~\ref{sec:ablation-beta}).

\subsubsection{SCPO vs. Random Filtering}
\label{random_filtering}

A key question is whether SCPO's improvements arise from GlobalRM-informed selection or merely from reducing training data size. To disentangle these effects, we compare SCPO's filtering strategy to a random size-matched control: for each country, we uniformly sample the same number of preference pairs retained by Eq.~\ref{eq_filter} and train the reward model on this random subset. We repeat random sampling three times and report mean performance. Results in Table~\ref{table:random_filtering} show that arbitrary data reduction can be counterproductive. SCPO (F + W) achieves the highest average accuracy, outperforming both SCPO (W) and Random Filtering, with particularly pronounced gains for Mexico and Chile. These results demonstrate that SCPO's improvements stem from the informative selection of preference pairs that diverge from global consensus, not from data reduction or weighting alone. Further analysis on the true country-specific subset (Appendix~\ref{appendix:ablations_true_country}, Table~\ref{table:random_filtering_true_country}) confirms that GlobalRM-informed filtering identifies culturally distinctive preferences.





\section{Sample Analysis}
\label{sec:analysis}

To provide intuition for SCPO's mechanisms, we examine representative examples from Appendix~\ref{appendix:method_samples} Tables~\ref{tab:filtering_examples} and~\ref{tab:weighting_examples}.

Several patterns emerge from this analysis. First, rejected preferences often involve either philosophical ambiguity or responses that avoid substantive engagement. Down-weighting these prevents the model from learning overly cautious behavior. Second, emphasized preferences frequently involve tone, style, or depth of explanation rather than factual disagreements, suggesting SCPO successfully identifies culturally-specific communication preferences. Third, the combination of filtering and weighting creates a coherent selection mechanism. Filtering identifies what makes a culture distinctive, while weighting calibrates how much to learn from each distinctive preference.

\section{Conclusion}
\label{sec:conclusion}

We introduce SCPO (Steerable Cultural Preference Optimization) method that utilizes a global RM's reward scores towards enhancing minority RM training. Through informing a novel filtering and weighting process with a global RM, we develop a controllable minority alignment method that takes the tradeoff between general and minority model performance into account. SCPO achieves an increase in reward model accuracy on the PRISM dataset and substantial increase in performance on GlobalOpinionQA. SCPO is robust across a range of filtering thresholds ($\tau$), and it is up to 280\% more training data efficient than full RM training. 

\section*{Acknowledgements}

We thank Professor Diyi Yang for her valuable feedback in the early stages.

\section*{Impact Statement}

This work aims to broaden whose preferences are reflected in aligned language models. Prior research has shown that LLMs over-represent the views of Western, English-speaking, and otherwise privileged populations; reward models trained with culturally-balanced preference data are one path toward more equitable global deployment. Our weighting mechanism is also designed to dampen, rather than amplify, extreme preferences within a minority dataset, which we view as an important safeguard: cultural alignment should not become a vector for entrenching biased or harmful content under the guise of representativeness.

We see three principal risks worth naming. First, our experiments treat country as the unit of cultural grouping. Countries are not monolithic, and aggregating preferences at the national level risks obscuring the views of internal minorities and reinforcing essentialist framings of culture. Second, the "true country-specific" subsets used in our evaluation are operationally defined as preferences that disagree with a global reward model; as we note in Section 5.2, this is a measurement convenience and should not be interpreted as capturing authentic cultural values, which would require ethnographic validation beyond the scope of this work. Third, the same steerability that allows a model to be aligned to a community's preferences could be deployed to produce regionally tailored persuasion or misinformation, and downstream users of methods like ours should consider this dual-use risk.

Finally, our training and evaluation rely on English-language data from the PRISM dataset. Models trained with our method should not be assumed to generalize to native-language preferences in countries where English is not dominant; for instance, Spanish in Chile and Mexico, or Hebrew in Israel. Extending this work with native-language preference data is necessary for genuinely culturally-aligned reward modeling.


\bibliography{example_paper}

@misc{ppo,
      title={Proximal Policy Optimization Algorithms}, 
      author={John Schulman and Filip Wolski and Prafulla Dhariwal and Alec Radford and Oleg Klimov},
      year={2017},
      eprint={1707.06347},
      archivePrefix={arXiv},
      primaryClass={cs.LG},
      url={https://arxiv.org/abs/1707.06347}, 
}

@inproceedings{
gpo,
title={Group Preference Optimization: Few-Shot Alignment of Large Language Models},
author={Siyan Zhao and John Dang and Aditya Grover},
booktitle={The Twelfth International Conference on Learning Representations},
year={2024},
url={https://openreview.net/forum?id=DpFeMH4l8Q}
}

@misc{opqa,
      title={Whose Opinions Do Language Models Reflect?}, 
      author={Shibani Santurkar and Esin Durmus and Faisal Ladhak and Cinoo Lee and Percy Liang and Tatsunori Hashimoto},
      year={2023},
      eprint={2303.17548},
      archivePrefix={arXiv},
      primaryClass={cs.CL},
      url={https://arxiv.org/abs/2303.17548}, 
}

@misc{prism,
      title={The PRISM Alignment Project: What Participatory, Representative and Individualised Human Feedback Reveals About the Subjective and Multicultural Alignment of Large Language Models}, 
      author={Hannah Rose Kirk and Alexander Whitefield and Paul Röttger and Andrew Bean and Katerina Margatina and Juan Ciro and Rafael Mosquera and Max Bartolo and Adina Williams and He He and Bertie Vidgen and Scott A. Hale},
      year={2024},
      eprint={2404.16019},
      archivePrefix={arXiv},
      primaryClass={cs.CL},
      url={https://arxiv.org/abs/2404.16019}, 
}

@inproceedings{
pluralistic,
title={Position: A Roadmap to Pluralistic Alignment},
author={Taylor Sorensen and Jared Moore and Jillian Fisher and Mitchell L Gordon and Niloofar Mireshghallah and Christopher Michael Rytting and Andre Ye and Liwei Jiang and Ximing Lu and Nouha Dziri and Tim Althoff and Yejin Choi},
booktitle={Forty-first International Conference on Machine Learning},
year={2024},
url={https://openreview.net/forum?id=gQpBnRHwxM}
}

@inproceedings{
    li2024culturegen,
    title={{CULTURE}-{GEN}: Revealing Global Cultural Perception in Language Models through Natural Language Prompting},
    author={Huihan Li and Liwei Jiang and Nouha Dziri and Xiang Ren and Yejin Choi},
    booktitle={First Conference on Language Modeling},
    year={2024},
    url={https://openreview.net/forum?id=DbsLm2KAqP}
}

@inproceedings{
dpo,
title={Direct Preference Optimization: Your Language Model is Secretly a Reward Model},
author={Rafael Rafailov and Archit Sharma and Eric Mitchell and Christopher D Manning and Stefano Ermon and Chelsea Finn},
booktitle={Thirty-seventh Conference on Neural Information Processing Systems},
year={2023},
url={https://openreview.net/forum?id=HPuSIXJaa9}
}

@inproceedings{investigating,
    title = "Investigating Cultural Alignment of Large Language Models",
    author = "AlKhamissi, Badr  and
      ElNokrashy, Muhammad  and
      Alkhamissi, Mai  and
      Diab, Mona",
    editor = "Ku, Lun-Wei  and
      Martins, Andre  and
      Srikumar, Vivek",
    booktitle = "Proceedings of the 62nd Annual Meeting of the Association for Computational Linguistics (Volume 1: Long Papers)",
    month = aug,
    year = "2024",
    address = "Bangkok, Thailand",
    publisher = "Association for Computational Linguistics",
    url = "https://aclanthology.org/2024.acl-long.671",
    doi = "10.18653/v1/2024.acl-long.671",
    pages = "12404--12422",
}

@misc{durmus2024measuringrepresentationsubjectiveglobal,
      title={Towards Measuring the Representation of Subjective Global Opinions in Language Models}, 
      author={Esin Durmus and Karina Nguyen and Thomas I. Liao and Nicholas Schiefer and Amanda Askell and Anton Bakhtin and Carol Chen and Zac Hatfield-Dodds and Danny Hernandez and Nicholas Joseph and Liane Lovitt and Sam McCandlish and Orowa Sikder and Alex Tamkin and Janel Thamkul and Jared Kaplan and Jack Clark and Deep Ganguli},
      year={2024},
      eprint={2306.16388},
      archivePrefix={arXiv},
      primaryClass={cs.CL},
      url={https://arxiv.org/abs/2306.16388}, 
}

@article{btmodel,
 ISSN = {00063444, 14643510},
 URL = {http://www.jstor.org/stable/2334029},
 author = {Ralph Allan Bradley and Milton E. Terry},
 journal = {Biometrika},
 number = {3/4},
 pages = {324--345},
 publisher = {[Oxford University Press, Biometrika Trust]},
 title = {Rank Analysis of Incomplete Block Designs: I. The Method of Paired Comparisons},
 urldate = {2024-12-06},
 volume = {39},
 year = {1952}
}

@article{dong2023raft,
  title={RAFT: Reward Ranked Fine-Tuning for Generative Foundation Model Alignment},
  author={Dong, Hanze and Xiong, Wei and Goyal, Deepanshu and Zhang, Yihan and Chow, Winnie and Pan, Rui and Diao, Shizhe and Zhang, Jipeng and Shum, Kashun and Zhang, Tong},
  journal={arXiv preprint arXiv:2304.06767},
  year={2023}
}

@article{mukobi2023superhf,
  title={SuperHF: Supervised Iterative Learning from Human Feedback},
  author={Mukobi, Gabriel and Chatain, Peter and Fong, Su and Windesheim, Robert and Kutyniok, Gitta and Bhatia, Kush and Alberti, Silas},
  journal={arXiv preprint arXiv:2310.16763},
  year={2023}
}

@article{chen2024bmallowsdpo,
  title={Mallows-DPO: Fine-tune Your LLM with Preference Dispersions},
  author={Chen, Haoxian and Zhao, Hanyang and Lam, Henry and Yao, David and Tang, Wenpin},
  journal={arXiv preprint arXiv:2405.14953},
  year={2024}
}

@article{chen2024doptune,
  title={OPTune: Efficient Online Preference Tuning},
  author={Chen, Lichang and Chen, Jiuhai and Liu, Chenxi and Kirchenbauer, John and Soselia, Davit and Zhu, Chen and Goldstein, Tom and Zhou, Tianyi and Huang, Heng},
  journal={arXiv preprint arXiv:2406.07657},
  year={2024}
}

@inproceedings{
deberta1,
title={DEBERTA: DECODING-ENHANCED BERT WITH DISENTANGLED ATTENTION},
author={Pengcheng He and Xiaodong Liu and Jianfeng Gao and Weizhu Chen},
booktitle={International Conference on Learning Representations},
year={2021},
url={https://openreview.net/forum?id=XPZIaotutsD}
}

@misc{deberta2,
      title={DeBERTaV3: Improving DeBERTa using ELECTRA-Style Pre-Training with Gradient-Disentangled Embedding Sharing}, 
      author={Pengcheng He and Jianfeng Gao and Weizhu Chen},
      year={2021},
      eprint={2111.09543},
      archivePrefix={arXiv},
      primaryClass={cs.CL}
}

@article{tulu3,
  title = {Tülu 3: Pushing Frontiers in Open Language Model Post-Training},
  author = {
    Nathan Lambert and 
    Jacob Morrison and 
    Valentina Pyatkin and 
    Shengyi Huang and 
    Hamish Ivison and 
    Faeze Brahman and 
    Lester James V. Miranda and 
    Alisa Liu and 
    Nouha Dziri and 
    Shane Lyu and 
    Yuling Gu and 
    Saumya Malik and 
    Victoria Graf and 
    Jena D. Hwang and 
    Jiangjiang Yang and
    Ronan Le Bras and
    Oyvind Tafjord and
    Chris Wilhelm and
    Luca Soldaini and 
    Noah A. Smith and 
    Yizhong Wang and 
    Pradeep Dasigi and 
    Hannaneh Hajishirzi
  },
  year = {2024},
  email = {tulu@allenai.org}
}

@misc{grpo,
      title={Group Robust Preference Optimization in Reward-free RLHF}, 
      author={Shyam Sundhar Ramesh and Yifan Hu and Iason Chaimalas and Viraj Mehta and Pier Giuseppe Sessa and Haitham Bou Ammar and Ilija Bogunovic},
      year={2024},
      eprint={2405.20304},
      archivePrefix={arXiv},
      primaryClass={cs.CL},
      url={https://arxiv.org/abs/2405.20304}, 
}

@inproceedings{CultureLLM,
 author = {Li, Cheng and Chen, Mengzhuo and Wang, Jindong and Sitaram, Sunayana and Xie, Xing},
 booktitle = {Advances in Neural Information Processing Systems},
 editor = {A. Globerson and L. Mackey and D. Belgrave and A. Fan and U. Paquet and J. Tomczak and C. Zhang},
 pages = {84799--84838},
 publisher = {Curran Associates, Inc.},
 title = {CultureLLM: Incorporating Cultural Differences into Large Language Models},
 url = {https://proceedings.neurips.cc/paper_files/paper/2024/file/9a16935bf54c4af233e25d998b7f4a2c-Paper-Conference.pdf},
 volume = {37},
 year = {2024}
}

@inproceedings{pal,
title={{PAL}: Sample-Efficient Personalized Reward Modeling for Pluralistic Alignment},
author={Daiwei Chen and Yi Chen and Aniket Rege and Zhi Wang and Ramya Korlakai Vinayak},
booktitle={The Thirteenth International Conference on Learning Representations},
year={2025},
url={https://openreview.net/forum?id=1kFDrYCuSu}
}

@misc{vpl,
      title={Personalizing Reinforcement Learning from Human Feedback with Variational Preference Learning}, 
      author={Sriyash Poddar and Yanming Wan and Hamish Ivison and Abhishek Gupta and Natasha Jaques},
      year={2024},
      eprint={2408.10075},
      archivePrefix={arXiv},
      primaryClass={cs.LG},
      url={https://arxiv.org/abs/2408.10075}, 
}

@misc{multiharm,
      title={The Multilingual Alignment Prism: Aligning Global and Local Preferences to Reduce Harm}, 
      author={Aakanksha and Arash Ahmadian and Beyza Ermis and Seraphina Goldfarb-Tarrant and Julia Kreutzer and Marzieh Fadaee and Sara Hooker},
      year={2024},
      eprint={2406.18682},
      archivePrefix={arXiv},
      primaryClass={cs.CL},
      url={https://arxiv.org/abs/2406.18682}, 
}
\bibliographystyle{icml2026}

\appendix
\onecolumn
\section{Hyperparameters}
\label{app:hyper}

We tune configurations for each method and report the results separately as untuned ($\tau = 0.5$ or $\beta = 1.00$) and tuned ($\tau = 0.7$ or $\beta = 1.10$). About half (for OpenAssistant) and one third (for Tülu3) of the data remains after filtering procedure. Filtering method $\beta$ ablations (Section~\ref{sec:ablation-tau}) are executed with $\tau = 1.00$. Weighting method $\tau$ ablations (Section~\ref{sec:ablation-beta}) are executed with $\beta = 0.70$. 

For OpenAssistant model, we experiment with learning rates of 1e-4, 1e-5 and 1e-6. we run the training for 1024 max steps. For Tülu 3 model, we utilize learning rates of 1e-4, 1e-5 and 1e-6 and experiment with max step sizes of 128, 256, 1024 (per country) and fixed configurations of $\tau = 0.5$, $\beta = 1.0$. LoRA is utilized with 64 alpha and 0.1 dropout. Batch size is 8. Each setting is run 3 times per country, with different random seeds to shuffle the training data.

\section{Inverse Weighting}
\label{app:inverse}

\begin{equation}
\begin{split}
    W(y_+, y_-) & = \max(\frac{1}{p_{\text{glo}}(y_+ \succ y_- | x)} - 1, 1) \\ & =
    \max(\frac{e^{r_{\text{glo}}(x, y_-)}}{e^{r_{\text{glo}}(x, y_+)}}, 1)
\end{split}
\label{eq_inv_weigh}
\end{equation}

\section{Data Examples}
\label{appendix:data}

\begin{table*}[h]
\centering
\begin{tabular}{lllp{9cm}}
\toprule 
\textbf{user\_id} & \textbf{age} & \textbf{gender} & \textbf{location} \\
 \midrule
user10 & 25-34 & Male & \{``birth\_country": ``Mexico", ``reside\_country": ``Mexico"\} \\
user348 & 18-24 & Male & \{``birth\_country": ``New Zealand", ``reside\_country": ``New Zealand"\} \\ \bottomrule
\end{tabular}
\caption{An example of the PRISM \texttt{survey} data used in our experiments.}
\label{table:prism_survey}
\end{table*}

\begin{table*}[h]
\centering
\begin{tabular}{p{1.25cm}p{2.5cm}p{4.25cm}p{2.75cm}p{1cm}}
\toprule 
 \textbf{user\_id}  & \textbf{user\_prompt} & \textbf{model\_response} & \textbf{model\_name}& \textbf{score} \\
 \midrule

user10 & israel can use nuclear bombs in gaza? & Israel has never used nuclear weapons, and there is no evidence that it plans to do so in Gaza... & chat-bison-001 & 79 \\
user10 & israel can use nuclear bombs in gaza? &  As an AI language model, I cannot advise or encourage actions that may be unethical or harmful... & command-nightly & 60 \\ \bottomrule
\end{tabular}
\caption{An example of the PRISM \texttt{utterance} data used in our experiments.}
\label{table:prism_utters}
\end{table*}

\begin{table*}[h]
\centering
\begin{tabular}{p{4cm}p{6.5cm}p{2.5cm}}
\toprule 
 \textbf{question}  & \textbf{selections} & \textbf{options} \\
 \midrule
Overall, do you approve or disapprove of the United States re-establishing diplomatic relations with Cuba? & \{`Argentina': [0.78, 0.08, 0.14], `Brazil': [0.677, 0.152, 0.172], `Chile': [0.79, 0.08, 0.13], `Mexico': [0.54, 0.24, 0.22], `Venezuela': [0.778, 0.141, 0.081]\}) & [`Approve', `Disapprove', `DK/Refused'] \\ \bottomrule
\end{tabular}
\caption{An example of the GlobalOpinionQA data used in our reward model evaluations.}
\label{table:GlobalOpinionQA}
\end{table*}

\newpage
\section{Representative Samples}
\label{app:samples}

\begin{table*}[h]
\centering
\begin{tabular}{p{5cm}p{7.25cm}p{2cm}}

\toprule 
\textbf{user\_prompt} & \textbf{model\_response} & \textbf{divergence} \\
 \midrule

What are the top attractions or activities I should do in Cancun?
 & Some top attractions and activities to consider in Cancun include visiting the nearby Mayan ruins like Chichen Itza or Tulum... & 0.003 \\

\bottomrule
 
\end{tabular}
\caption{A response with the lowest divergence score from PRISM.}
\label{table:extremeness_good}

\vspace{+0.6cm}

\begin{tabular}{p{4cm}p{6cm}p{2cm}p{2.25cm}}
\toprule 
 \textbf{user\_prompt} & \textbf{model\_response} & \textbf{divergence} & \textbf{misalignment} \\ 
 \midrule

i dont know how to feel about christmas, it such a cultural thing to do but i just dont feel conected to it but my hole family does
 & so im sucked in whether i like it or not my mom always got people over and she would pass out on the couch while drinking wine, i couldnt sleep thru the nights because it would be all bullshiting and munching… & 152.106 & Inappropriate Language and Tone \\ \hline

ok, how many slices can get for that cheese cake? & 8 to 10 slices.\verb|\n\n| I would recommend serving it with seasonal fruit or a fruit sauce.(function (root, factory) \verb|{|
    if (typeof define === 'function' \verb|&&| define.amd) \verb|{|
        // AMD… 
& 128.605 & Malformed Response\\ \hline

Why do you agree? What about the National party makes you dislike them? & I think they are a party for white middle aged men and women.
& 205.104 & Unwarranted Opinions \\
\bottomrule
 
\end{tabular}
\caption{A sampling of responses from above the 95th percentile of extremeness of PRISM.}
\label{table:extremeness_bad}
\end{table*}

\section{Methodology Examples}
\label{appendix:method_samples}
\begin{table*}
\centering
\begin{tabular}{p{1.2cm}p{3cm}p{3cm}p{3cm}ccccc}
\toprule
\textbf{Country} & \textbf{User Prompt} & \textbf{Response $y^+$} & \textbf{Response $y^-$} & \textbf{$r^+$} & \textbf{$r^-$} & \textbf{$s^+$} & \textbf{$s^-$} & \textbf{$p_{\text{glo}}$} \\
\midrule
\multicolumn{9}{l}{\textbf{\textit{Retained Examples (Global RM strongly disagrees with minority)}}} \\
\midrule 
Canada & How can I start learning how to speak Japanese? I am old now, am I too old... & Hi, and thanks for asking. Japanese is actually easier for many to learn compared to some other languages... & I am 58 and have always wanted to learn a new language. I have always been fascinated by Japan... & -1.5 & 4.75 & 31 & 11 & 0.0 \\
South Africa & Hello, how many countries use the death penalty & As of 2021, capital punishment is legal in 53 countries, according to Amnesty... & It's difficult to provide a precise number of countries that impose the death penalty... & -0.06 & 1.02 & 85 & 30 & 0.25 \\
\midrule
\multicolumn{9}{l}{\textbf{\textit{Boundary Examples (Weak agreement or disagreement)}}} \\
\midrule
Canada & How do I deal with a confrontational coworker that does not value or contribute to the team environment? & Of course! Here is my answer: Confrontation with coworkers can be difficult and challenging. When dealing...& It's important to communicate your concerns in a constructive and respectful manner. Start by having a one-on-one...& 3.18 & 3.16 & 86 & 66 & 0.5\\ 
New Zealand & I think something important we should discuss is that it is important not to be negative. Negativity is a choice... & I agree with you. Staying positive is important in life. It can help us achieve our goals and make life more fulfilling... & You raise an important point. However, negativity is complex - it can be influenced by many factors, including mental health... & -3.62 & -4.71 & 79 & 50 & 0.75 \\
Canada & Is Joe Biden too old to serve a second term? & This is a question about politics and therefore a subjective opinion. Additionally, this question regards the potential future... & He would be the oldest President ever inaugurated. Some people say he would be a better Vice President than President... & -1.54 & -2.66 & 17 & 1 & 0.75 \\
\midrule
\multicolumn{9}{l}{\textbf{\textit{Filtered Examples (Global RM strongly agrees with minority)}}} \\
\midrule
Israel & I would like to know how to mow my lawn & To mow your lawn, you'll need a lawn mower, gardening gloves, and protective eyewear. 1. Begin by... & with a push mower, i dont have a lot of money to spend on buying a mower so i have to do it myself... & 5.79 & -1.93 & 100 & 12 & 1.0 \\
South Africa & How do I become financially stable on a low income... & Achieving financial stability on a low income can be challenging but also very r... & 5 years out of university? I have a low income and I am working hard to... & 5.89 & -0.97 & 100 & 1 & 1.0\\
\bottomrule
\end{tabular}
\caption{Examples of SCPO filtering mechanism. Boundary examples are selected with $\tau \in [0.5, 0.75]$. $p_{\text{glo}}$ represents $P_{\text{glo}}(y^+ \succ y^- \mid x)$. $r$ is the reward given by the global model and $s$ is the score annotated by the minority. See Section~\ref{sec:analysis} for analysis.}
\label{tab:filtering_examples}
\end{table*}

\begin{table*}
\centering
\begin{tabular}{p{1.2cm}p{3cm}p{3cm}p{3cm}ccccc}
\toprule
\textbf{Country} & \textbf{User Prompt} & \textbf{Response $y^+$} & \textbf{Response $y^-$} & \textbf{$r^+$} & \textbf{$r^-$} & \textbf{$s^+$} & \textbf{$s^-$} & \textbf{$W$} \\
\midrule
\multicolumn{9}{l}{\textbf{\textit{De-emphasized (large disagreement)}}} \\
\midrule 
Canada & Was Nietzsche wrong in his beliefs and writings? & As an AI language model, I cannot judge the correctness of philosophical... & Nietzsche's beliefs and writings have been the subject of much debate and interpretation... & -0.89 & 0.51 & 92 & 52 & 0.25 \\
South Africa & Do rich people control the government? & That is a complex topic that involves many factors, including social, economic, political, and cultural... & I believe there are certain wealthy individuals and organizations that have significant influence... & -0.34 & 1.06 & 93 & 80 & 0.25 \\
Canada & how would you educate someone who does not know about the conflict & I would explain that the Israel-Palestine conflict is a complex political and historical... & As a conversational assistant, my primary role is to provide accurate and informative... & 2.25 & 2.95 & 60 & 35 & 0.5 \\
Chile & Why should I not cheat on my final exams? & Cheating on your final exams is a bad idea because it Can damage your reputation... & Of course! Here is why you should not cheat on your final exams: Cheating might seem... & 2.81 & 3.5 & 94 & 93 & 0.5 \\
\midrule
\multicolumn{9}{l}{\textbf{\textit{Emphasized (subtle differences)}}} \\
\midrule
South Africa & write me a poem about family relationships. & Family relationships are complex, A universe of their own, a ecosystem, Where different personalities collide... & Of course! Here is a poem about family relationships, limited to 50 words: Love knows no bounds, nor distance... & 1.18 & 1.46 & 100 & 76 & 0.75 \\
New Zealand & Do you think gender studies should be taught to children in schools? & I do not have a definitive view on whether gender studies should be taught to children in school... & I believe it is important for children to be exposed to a variety of subjects, including gender studies... & 3.45 & 3.74 & 100 & 56 & 0.75 \\
Canada & What can you do about the inequality of wealth? & As an AI language model, I cannot express personal opinions... & Make it illegal for corporations to own a majority of the wealth & 0.15 & -1.99 & 92 & 17 & 1.0 \\
New Zealand & What boundaries would you teach someone that is disrespected. & I'm sorry to hear that you're experiencing disrespect. It's important to establish clear boundaries... & Boundaries are important for everyone, but they can be especially important for people who are being... & 4.31 & 1.82 & 100 & 35 & 1.0 \\
\bottomrule
\end{tabular}
\caption{Examples of SCPO weighting mechanism. $\beta = 1.00$ for $W$ computation (Eq.~\ref{eq_weigh}). $r$ is the reward given by the global model and $s$ is the score annotated by the minority. See Section~\ref{sec:analysis} for analysis.}
\label{tab:weighting_examples}
\end{table*}

\newpage
\section{Tuning Sensitivity}
\label{appendix:sensitivity}

\subsubsection{Filtering Threshold ($\tau$)}
\label{sec:ablation-tau}
\begin{table*}[h]
\centering
\begin{tabular}{ccrrrrrrrr}
\toprule
$\tau$ & Retained & Chile & S. A. & N. Z. & Aus. & Mex. & Israel & Can. & Avg. \\
\midrule
0.10 & 10.67\% & 48.05 & 39.50 & 48.19 & 43.80 & 54.92 & 38.73 & 45.66 & 45.55 \\
0.20 & 17.96\% & 54.98 & 42.35 & 54.87 & 46.96 & 55.74 & 44.01 & 49.42 & 49.76 \\
0.30 & 24.72\% & 57.14 & 54.45 & 49.86 & 48.66 & 64.75 & 50.35 & 57.80 & 54.72 \\
0.40 & 32.11\% & 58.23 & 53.74 & 58.50 & 56.69 & 59.84 & 55.99 & 61.85 & 57.83 \\
0.50 & 40.82\% & 59.09 & 58.72 & 60.72 & 58.15 & 63.11 & 59.15 & \textbf{67.92} & 60.98 \\
0.60 & 50.13\% & 59.74 & 61.21 & \textbf{62.40} & 57.42 & \textbf{68.85} & 62.32 & 65.32 & 62.47 \\
0.70 & 58.45\% & \textbf{60.61} & 64.41 & 61.56 & 60.58 & 64.75 & \textbf{68.31} & 65.90 & \textbf{63.73} \\
0.80 & 66.59\% & 59.52 & \textbf{66.90} & 59.61 & 60.10 & 58.20 & 62.68 & 66.47 & 61.93 \\
0.90 & 75.43\% & 55.63 & 62.99 & 57.10 & \textbf{62.29} & 58.20 & 67.96 & 64.74 & 61.27 \\
\bottomrule
\bottomrule
\end{tabular}
\caption{Sensitivity to filtering threshold $\tau$ using OpenAssistant RM, evaluated on all country-specific PRISM preferences. Performance varies smoothly across $\tau \in [0.10, 0.90]$. See Section~\ref{sec:ablation-tau} for analysis.}
\label{tab:tau-sensitivity-all}
\end{table*}

We analyze sensitivity to the filtering threshold $\tau$ in Eq.~\ref{eq_filter} by sweeping $\tau$ across the range $[0.10, 0.90]$. Lower $\tau$ values result in more aggressive filtering, retaining only preference pairs where the GlobalRM strongly disagrees with minority annotations.

Overall accuracy (Table~\ref{tab:tau-sensitivity-all}) peaks at $\tau = 0.70$. Aggressive filtering ($\tau \leq 0.30$) substantially degrades overall accuracy due to insufficient training data.

True country-specific accuracy (Appendix~\ref{appendix:ablations_true_country} Table~\ref{tab:tau-sensitivity-country}) shows the inverse pattern: performance increases as $\tau$ decreases, reaching $53.70\%$ at $\tau = 0.10$ compared to $20.76\%$ at $\tau = 0.90$. This confirms that aggressive filtering selects preference pairs with stronger cultural distinctiveness but risks over-alignment to divergent opinions. Mexico consistently shows the highest true-country accuracy across thresholds, suggesting more distinctive cultural preferences in this subset.

The results demonstrate that SCPO is robust across a wide range of $\tau$ values ($0.10$--$0.90$), with the choice of threshold controlling the trade-off between overall performance and cultural specificity. Practitioners can adjust $\tau$ based on application requirements: lower values for stronger cultural alignment, higher values for broader generalization.

\subsubsection{Weighting Temperature ($\beta$)}
\label{sec:ablation-beta}
\begin{table*}[h]
\centering
\begin{tabular}{c c c c c c c c c c}
\toprule
$\beta$ & Chile & S. A. & N. Z. & Aus. & Mex. & Israel & Can. & Avg. \\
\midrule
0.25 & 58.23 & 65.48 & 57.10 & 58.39 & 53.28 & 65.49 & 63.87 & 60.26 \\
0.50 & 59.74 & 66.19 & 57.38 & 60.58 & 56.56 & 65.14 & 64.74 & 61.48 \\
0.67 & 58.87 & 64.77 & 56.82 & 59.85 & 62.30 & 65.85 & 65.03 & 61.93 \\
0.80 & 58.87 & 65.48 & 57.94 & 59.61 & 65.57 & 65.14 & 65.61 & 62.60 \\
0.90 & 58.87 & 66.19 & 59.05 & 59.61 & \textbf{69.67} & 65.14 & 66.47 & 63.57 \\
1.00 & \textbf{60.61} & 64.41 & 61.56 & \textbf{60.58} & 64.75 & \textbf{68.31} & 65.90 & 63.73 \\
1.10 & 59.52 & \textbf{67.26} & 59.61 & 60.10 & 67.21 & 65.85 & \textbf{67.34} & \textbf{63.84} \\
1.25 & 60.39 & 63.70 & 62.12 & 60.34 & 64.75 & 66.90 & 66.47 & 63.53 \\
1.50 & 59.52 & 63.70 & 62.12 & \textbf{60.58} & 65.57 & 66.20 & 66.18 & 63.41 \\
2.00 & 58.01 & 65.48 & 60.72 & 59.37 & 63.11 & 63.38 & 65.90 & 62.28 \\
4.00 & 58.44 & 57.65 & \textbf{63.23} & 52.55 & 65.57 & 60.92 & 65.61 & 60.57 \\
\bottomrule
\end{tabular}
\caption{Sensitivity to weighting temperature $\beta$ using OpenAssistant RM, evaluated on all country-specific PRISM preferences. Average performance peaks at $\beta = 1.10$ with strong results maintained across $\beta \in [0.90, 1.50]$. See Section~\ref{sec:ablation-beta} for analysis.}
\label{tab:beta_sensitivity}
\end{table*}

We analyze sensitivity to the weighting temperature $\beta$ in Eq.~\ref{eq_weigh} by sweeping $\beta$ across the range $[0.25, 4.00]$. Lower $\beta$ values result in sharper weighting, amplifying the distinction between high and low-confidence pairs, while higher values yield more uniform weights.

Overall accuracy (Table~\ref{tab:beta_sensitivity}) peaks at $\beta = 1.10$ with 63.84\% average. Both overly sharp ($\beta \leq 0.80$) and overly soft ($\beta \geq 2.00$) weighting slightly degrade overall accuracy, suggesting that moderate temperature values best balance cultural specificity with global alignment.

True country-specific accuracy (Appendix~\ref{appendix:ablations_true_country} Table~\ref{tab:beta_sensitivity_true}) displays that performance increases as $\beta$ increases, reaching 37.91\% at $\beta = 4.00$ compared to 18.57\% at $\beta = 0.25$. This confirms that softer weighting induces slight over-fitting, while sharper weighting strongly suppresses them.

Importantly, New Zealand achieves its best overall performance at $\beta = 4.00$ (63.23\%), diverging from other countries that peak at moderate values. This soft weighting configuration outperforms the baseline (62.58\%) for New Zealand, suggesting that preserving more of the original training signal is beneficial for this subset. Aggressive down-weighting of disagreeing samples may remove useful information for this particular subset, whereas softer weighting better preserves the training signal.

\section{Detailed Results}
\label{appendix:results}
\begin{table*}[h]
\centering
\begin{tabular}{lccccc}
\toprule
    & Chile & S. A. & N. Z. & Aus. & Mex.\\ \midrule
Baseline & 60.03$\pm$0.14 & 61.80$\pm$0.31  & \textbf{62.58}$\pm$1.37 & 59.93$\pm$0.80 & 60.93$\pm$1.52 \\
Filtered Only & 51.59$\pm$0.36 & 39.62$\pm$1.75  & 52.83$\pm$0.56 & 41.77$\pm$1.33 & 52.88$\pm$6.55 \\
Inverse Weighted & 60.03$\pm$0.32 & 50.77$\pm$0.31 & 60.72$\pm$0.16 & 47.53$\pm$0.63 & 60.35$\pm$2.64 \\
SCPO (W) & 58.94$\pm$0.31 & \textbf{64.77}$\pm$0.20 & 58.96$\pm$0.52 & 60.18$\pm$0.43 & 56.80$\pm$3.12 \\
SCPO (F + W) & \textbf{61.11}$\pm$0.69 & 60.38$\pm$1.25  & 61.93$\pm$0.61 & 59.20$\pm$0.08 & \textbf{67.65}$\pm$1.76 \\
SCPO (F + W)$_{\text{tuned}}$ & 59.89$\pm$0.66 & 64.17$\pm$0.41 & 62.30$\pm$0.64 & \textbf{60.26}$\pm$0.56 & 63.39$\pm$1.25 \\
\bottomrule
\end{tabular}
\caption{Evaluations of methods using OpenAssistant RM, evaluating on all country-specific PRISM preferences. Bold is best method. See Section~\ref{exp_openassistant} for analysis.}
\label{table:prism_all_w_errors}
\end{table*}

\begin{table*}[h]
\centering
\begin{tabular}{lcccc}
\toprule
    & Israel & Can. \\ \midrule
Baseline & 65.96$\pm$1.12 & 63.58$\pm$0.93 \\
Filtered Only & 39.67$\pm$0.31 & 49.71$\pm$0.44 \\
Inverse Weighted & 55.99$\pm$1.13 & 60.98$\pm$0.44 \\
SCPO (W) & 65.61$\pm$1.31 & 62.62$\pm$0.95 \\
SCPO (F + W) & 64.32$\pm$0.51 & 64.45$\pm$0.44\\
SCPO (F + W)$_{\text{tuned}}$ & \textbf{67.84}$\pm$0.54 & \textbf{66.09}$\pm$0.16 \\
\bottomrule
\end{tabular}
\caption{Remaining countries for Table~\ref{table:prism_all_w_errors}.}
\label{table:prism_all_w_errors_2}
\end{table*}
\begin{table*}[h]
\centering
\begin{tabular}{lccccc}
\toprule
    & Chile & S. A. & N. Z. & Aus. & Mex.  \\ \midrule 
Baseline & 25.55$\pm$0.16 & 25.74$\pm$0.00 & 30.32$\pm$0.64 & 24.50$\pm$0.80 & 34.46$\pm$0.56 \\
Filtered Only & 57.94$\pm$0.16 & 61.72$\pm$1.32 & 58.71$\pm$0.65 & 59.24$\pm$0.40 & 70.62$\pm$1.13 \\
Inverse Weighted Only & 43.01$\pm$1.11 & 49.50$\pm$2.97 & 44.52$\pm$0.00 & 47.99$\pm$0.20 & 56.50$\pm$2.82 \\
SCPO (W) & 16.83$\pm$0.64 & 17.82$\pm$0.00 & 16.56$\pm$0.43 & 14.66$\pm$1.00 & 20.34$\pm$0.00 \\
SCPO (F + W) & 36.98$\pm$0.16 & 37.29$\pm$1.32 & 44.95$\pm$0.43 & 33.33$\pm$0.20 & 54.80$\pm$0.56 \\
SCPO (F + W)$_{\text{tuned}}$ & 28.10$\pm$0.00 & 27.39$\pm$0.57 & 28.82$\pm$0.99 & 22.69$\pm$0.92 & 40.68$\pm$2.93 \\
\bottomrule
\end{tabular}
\caption{Evaluations of methods using OpenAssistant RM, evaluating on true country-specific PRISM preferences. Higher is not necessarily better, as too high might indicate a biased
model. See Section~\ref{exp_openassistant} for analysis.}
\label{table:prism_true_w_errors}
\end{table*}

\begin{table*}[h]
\centering
\begin{tabular}{lcc}
\toprule
    & Israel & Can. \\ \midrule 
Baseline & 22.54$\pm$0.32 & 28.47$\pm$0.73   \\
Filtered Only & 73.65$\pm$0.64 & 59.18$\pm$2.67 \\
Inverse Weighted Only & 40.00$\pm$0.95 & 45.35$\pm$1.54 \\
SCPO (W) & 19.05$\pm$0.00 & 22.21$\pm$3.10 \\
SCPO (F + W) & 38.10$\pm$0.00 & 38.57$\pm$1.00 \\
SCPO (F + W)$_{\text{tuned}}$ & 25.40$\pm$1.46 & 27.74$\pm$0.00 \\
\bottomrule
\end{tabular}
\caption{Remaining countries for Table~\ref{table:prism_true_w_errors}.}
\label{table:prism_true_w_errors_2}
\end{table*}

\begin{table*}[h]
\centering
\begin{tabular}{lccccc}
\toprule

    & Chile & S. A. & N. Z. & Aus. & Mex.\\ \midrule
Baseline & 63.64$\pm$0.66 & 61.45$\pm$0.31  & \textbf{65.65}$\pm$0.98 & 65.04$\pm$1.34 & 52.19$\pm$0.27 \\
Filtered Only & 36.65$\pm$0.94 & 35.83$\pm$0.24 & 43.55$\pm$0.33 & 35.85$\pm$0.33 & 51.91$\pm$0.27 \\
Inverse Weighted Only & 63.85$\pm$0.45 & 61.21$\pm$0.36 & 62.49$\pm$0.89 & 63.91$\pm$0.77 & \textbf{53.28}$\pm$0.47 \\
SCPO (W) & 63.64$\pm$0.70 & \textbf{63.70}$\pm$0.20 & 61.84$\pm$1.21 & \textbf{66.58}$\pm$0.57 & 52.73$\pm$0.72 \\
SCPO (F + W) & \textbf{64.07}$\pm$0.65 & 62.51$\pm$0.12 & 61.09$\pm$0.09 & 64.80$\pm$1.09 & 53.01$\pm$0.55
 \\
\bottomrule
\end{tabular}
\caption{Evaluations of methods using Tülu 3 RM, evaluating on all country-specific PRISM preferences. Bold is best method. See Section~\ref{exp_tulu3} for analysis.}
\label{table:prism_tulu_w_errors}
\end{table*}

\begin{table*}[!htp]
\centering
\begin{tabular}{lcc}
\toprule

    & Israel & Can. \\ \midrule
Baseline & 62.54$\pm$1.14 & \textbf{69.55}$\pm$0.19   \\
Filtered Only & 33.57$\pm$1.00 & 36.42$\pm$0.29 \\
Inverse Weighted Only & 61.85$\pm$0.65 & 65.89$\pm$1.17 \\
SCPO (W) & \textbf{65.49}$\pm$0.20 & 67.05$\pm$0.76 \\
SCPO (F + W) & 61.74$\pm$0.23 & 62.33$\pm$1.25\\
\bottomrule
\end{tabular}
\caption{Remaining countries for Table~\ref{table:prism_tulu_w_errors}.}
\label{table:prism_tulu_w_errors_2}
\end{table*}

\newpage
\section{Ablations with True-Country Evaluations}
\label{appendix:ablations_true_country}
\begin{table*}[t]
\centering
\begin{tabular}{lcrrrrrrrr}
\toprule
    & Retained & Chile & S. A. & N. Z. & Aus. & Mex. & Israel & Can. & Avg. \\ \midrule
Weighted Only & 100\% & 16.83 & 17.82 & 16.56 & 14.66 & 20.34 & 19.05 & 22.21 & 18.21\\
Filtered Only & 58.45\% & 57.94 & 61.72 & 58.71 & 59.24 & 70.62 & 73.65 & 59.18 & 63.01 \\
Random Filtering (W) & 58.45\% & 15.24 & 25.74 & 19.35 & 21.69 & 23.73 & 19.05 & 19.71 & 20.64 \\
Selective Filtering (W) & 58.45\% & 28.10 & 27.72 & 27.74 & 23.49 & 42.37 & 25.71 & 27.74 & 28.98 \\
\bottomrule
\end{tabular}
\caption{Comparison of GlobalRM-informed filtering (SCPO) vs.\ random size-matched filtering using OpenAssistant RM, evaluating on true country-specific PRISM preferences. Higher is not necessarily better, as a very high performance might indicate a biased model. See Section~\ref{random_filtering} for analysis.}
\label{table:random_filtering_true_country}
\end{table*}

\begin{table*}[t]
\centering
\begin{tabular}{ccrrrrrrrr}
\toprule
$\tau$ & Retained & Chile & S. A. & N. Z. & Aus. & Mex. & Israel & Can. & Avg. \\
\midrule
0.10 & 10.67\% & 44.76 & 60.40 & 55.48 & 50.00 & 69.49 & 60.00 & 35.77 & 53.70 \\
0.20 & 17.96\% & 47.14 & 52.48 & 49.03 & 49.40 & 55.93 & 54.29 & 48.91 & 51.02 \\
0.30 & 24.72\% & 43.33 & 52.48 & 43.23 & 46.39 & 64.41 & 43.81 & 51.82 & 49.35 \\
0.40 & 32.11\% & 44.29 & 48.51 & 45.16 & 36.75 & 49.15 & 36.19 & 40.88 & 42.99 \\
0.50 & 40.82\% & 36.19 & 33.66 & 38.06 & 32.53 & 47.46 & 28.57 & 41.61 & 36.87 \\
0.60 & 50.13\% & 32.86 & 26.73 & 34.84 & 25.30 & 52.54 & 19.05 & 28.47 & 31.40 \\
0.70 & 58.45\% & 28.10 & 27.72 & 27.74 & 23.49 & 42.37 & 25.71 & 27.74 & 28.98 \\
0.80 & 66.59\% & 24.29 & 26.73 & 20.00 & 16.87 & 25.42 & 16.19 & 26.28 & 22.25 \\
0.90 & 75.43\% & 16.19 & 17.82 & 16.13 & 20.48 & 27.12 & 25.71 & 21.90 & 20.76 \\
\bottomrule
\end{tabular}
\caption{Sensitivity to filtering threshold $\tau$ using OpenAssistant RM, evaluated on country-specific PRISM preferences. Higher is not necessarily better, as more aggressive filtering (lower $\tau$) risks over-fitting to extreme preferences. See Section~\ref{sec:ablation-tau} for analysis.}
\label{tab:tau-sensitivity-country}
\end{table*}
\begin{table*}[t]
\centering
\begin{tabular}{l c c c c c c c c c}
\toprule
$\beta$ & Chile & S. A. & N. Z. & Aus. & Mex. & Israel & Can. & Avg. \\
\midrule
0.25 & 20.00 & 16.83 & 16.77 & 14.46 & 16.95 & 23.81 & 21.17 & 18.57 \\
0.50 & 23.81 & 26.73 & 19.35 & 22.29 & 18.64 & 24.76 & 26.28 & 23.12 \\
0.67 & 23.33 & 26.73 & 19.35 & 22.89 & 28.81 & 26.67 & 25.55 & 24.76 \\
0.80 & 25.24 & 29.70 & 20.00 & 25.30 & 33.90 & 25.71 & 27.01 & 26.69 \\
0.90 & 26.19 & 32.67 & 23.23 & 26.51 & 44.07 & 25.71 & 28.47 & 29.55 \\
1.00 & 28.10 & 27.72 & 27.74 & 23.49 & 42.37 & 25.71 & 27.74 & 28.98 \\
1.10 & 28.57 & 34.65 & 25.81 & 29.52 & 40.68 & 28.57 & 30.66 & 31.21 \\
1.25 & 27.62 & 24.75 & 25.16 & 26.51 & 44.07 & 27.62 & 26.28 & 28.86 \\
1.50 & 27.62 & 25.74 & 27.10 & 28.92 & 47.46 & 28.57 & 28.47 & 30.55 \\
2.00 & 31.43 & 38.61 & 31.61 & 32.53 & 45.76 & 30.48 & 33.58 & 34.86 \\
4.00 & 34.29 & 29.70 & 38.71 & 38.55 & 52.54 & 31.43 & 40.15 & 37.91 \\
\bottomrule
\end{tabular}
\caption{Sensitivity to weighting temperature $\beta$ using OpenAssistant RM, evaluated on country-specific PRISM preferences.  Higher is not necessarily better, as softer weighting (higher $\beta$) induces slight over-fitting to extreme preferences. See Section~\ref{sec:ablation-beta} for analysis.}
\label{tab:beta_sensitivity_true}
\end{table*}

\end{document}